\documentclass{amia}
\usepackage{amsmath}
\usepackage{booktabs}
\usepackage{xcolor}
\usepackage{graphicx}
\usepackage{tikz}
\usepackage{hyperref}
\usepackage{multirow}
\usepackage{subcaption}
\usepackage{pgfplots}
\pgfplotsset{compat=1.18}
\usepackage{lipsum} %Remove if not needed
\begin{document}

\title{Relation Extraction with Instance-Adapted Predicate Descriptions}

\author{Yuhang Jiang, MS and  Ramakanth Kavuluru, PhD }

\institutes{
Division of Biomedical Informatics, Department of Internal Medicine\\
    University of Kentucky, Lexington, KY, USA
}

\maketitle

\section*{Abstract}

\textit{Relation extraction (RE) is a standard information extraction task playing a major role in downstream applications such as knowledge discovery and question answering. Although decoder-only large language models are excelling in generative tasks, smaller encoder models are still the go to architecture for RE. In this paper, we revisit fine-tuning such smaller models using   a novel dual-encoder architecture with a  joint contrastive and cross-entropy loss. Unlike previous methods that employ a fixed linear layer for predicate representations, our approach uses a second encoder to compute instance-specific predicate representations by infusing them with real entity spans from corresponding input instances. We conducted experiments on two biomedical RE datasets and two general domain datasets. Our approach achieved  $F_1$ score improvements ranging from 1\% to 2\% over  state-of-the-art  methods with a simple but elegant formulation. Ablation studies justify the  importance of various components built into the proposed architecture. }

\section*{Introduction}
\vspace{-1mm}
Relation extraction (RE) is a basic task in natural language processing (NLP), especially in applied domains such as biomedicine and healthcare where relations among biomedical entities drive disease and treatment processes. A relation typically connects a subject entity and an object entity via a predicate (or relation type) as in (\textit{tamoxifen}, \texttt{treats}, \textit{breast cancer}). The goal is to extract such relations from natural language inputs,  with at times the added goal of normalizing the entity spans to standardized vocabularies. Having a database of relations pertinent to a domain of interest can enable knowledge discovery and question answering.

\textbf{\textit{Relation extraction trends}: }
Early RE efforts focused on rule-based systems, kernel methods, and shortest path algorithms~\cite{riloff1993automatically, zelenko-etal-2002-kernel, bunescu-mooney-2005-shortest}. As the field evolved, methods shifted to purely supervised models with labeled data. An initial approach was to use n-gram features leveraging dependency paths between the subject and object entity spans~\cite{kambhatla2004combining}.
Subsequently neural embeddings, convolutional~\cite{nguyen2015relation} and recurrent architectures~\cite{miwa2016end} and their combinations~\cite{vu2016combining} enhanced with attention mechanisms~\cite{guo2019attention} became popular. Since transformers were invented, the BERT architecture and its variants became popular for named entity recognition (NER) and RE efforts~\cite{lin2019bert,joshi2020spanbert}. 

\begin{figure}[h]
    \centering
    \includegraphics[scale=0.4]{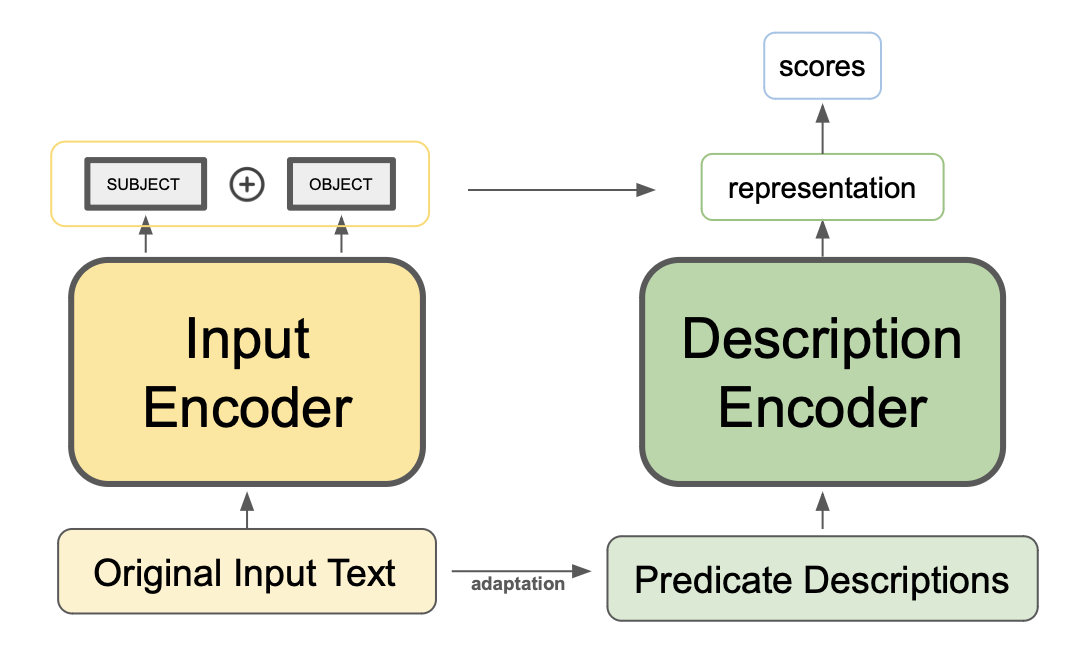}
    \caption{A dual-encoder architecture with instance adapted predicate descriptions for relation extraction.}
    \label{fig:intro-fig}
\end{figure}

There was a general consensus that joint end-to-end modeling (where entities and relations are extracted together in a single model) was better over pipeline based approaches (where an NER model and a separate RE model are stitched to form a pipeline). However, Zhong and Chen~\cite{zhong-chen-2021-frustratingly} challenged that paradigm and showed that pipelines can still be superior with a clever marker based representation for entities. So pipelines are going through a revival and it is still worthwhile to build separate high quality models for NER and RE and join them in the end. Here, the RE component assumes the entities are already spotted. In this paper, we focus on this RE component that identifies relations between pre-spotted entities provided as part of the input.

We realize that decoder-only (autoregressive) large language models (LLMs) have become quite popular for general NLP tasks. While they clearly excel at generative tasks (e.g., summarization) and zero and few-shot situations for RE~\cite{li2023revisiting}, there is scarce evidence (if any) that they perform on par with encoder models when ample training data is available; their use has been mostly limited to data augmentation to enhance training dataset with synthetic examples while the eventual  model to be trained is still a BERT variant or a encoder-decoder model such as T5~\cite{wadhwa2023revisiting}. 

\textbf{\textit{High level idea of our method}: }
\label{sec-high}
Here we set out to improve  relation classification abilities of encoder models. 
The default approach to RE once the entities are spotted is to derive entity (span) representations using the encoder and merging them in some way (typically, via concatenation) to derive softmax probability estimates for all predicates including the \texttt{NULL} (no relation) label. This corresponds to the left  block of Figure~\ref{fig:intro-fig} (note that softmax layer is not shown in the figure). 

We propose to use predicate descriptions or definitions as an auxiliary signal. Most RE datasets/tasks, especially in biomedicine, have definitions of what a predicate is supposed to encode in relations that use it. For example, the US National Library of Medicine's semantic network\footnote{\url{https://lhncbc.nlm.nih.gov/semanticnetwork/}} has the following official definition for the \texttt{TREATS} predicate: ``Applies a remedy with the object of effecting a cure or managing a condition.'' This could be seen as a canonical way of describing a treatment relation although people could discuss it in myriad ways in natural language. The high level idea is to first derive an instance-adapted predicate description by \textit{instantiating} the canonical predicate definition with entity spans from input text. Next, compare this description with the input text and pick the predicate whose instance-adapted description \textit{matches} the most with the input. For example, consider the input sentence with italicized entity spans: ``\textit{Tamoxifen} is the most common endocrine therapy administered worldwide to women with hormone receptor-positive metastatic \textit{breast cancer}.''
The \texttt{treats} predicate associated description for this instance is: ``Applies a \textit{tamoxifen} remedy with the object of effecting a cure or managing a \textit{breast cancer} condition.'' It is straightforward to see this description semantically matches better with the input sentence compared with descriptions of other predicates (e.g., \texttt{CAUSES}). We carry this out using a dual encoder architecture as shown in Figure~\ref{fig:intro-fig} where the left block encodes the input instance and the right block encodes the instance-specific predicate descriptions. 
In the rest of the paper, we formalize this intuition and evaluate the resulting method with four different datasets. We show $F_1$ score improvements ranging from 1\% to 2.1\% compared to prior best methods. The datasets we used are all public and our code is available at: \url{https://github.com/bionlproc/Instance-Adapted-RE}.

\section*{Methods}
\vspace{-1mm}
We first elaborate on the dual encoder architecture being proposed in this paper. Formally, for any input text containing mentions of entity spans say constituting set $E$, the goal is to determine a predicate $r \in R$ for each possible pair $(e_s, e_o) \in E \times E$, where $R$ includes the generally most frequent \texttt{NULL} predicate. 
As indicated in the Introduction, we have two encoders, one for the original input text and one for the input adapted predicate descriptions. We first discuss the input text encoder.

\textbf{\textit{Input representation}: }
It is important to note we are representing the input text along with an entity pair $(e_s, e_o) \in E \times E$ to classify if they participate in a relation as asserted in the input. Since the eventual classification is dependent on the particular pair of entities, the representation is a linear projection of the entity embeddings from the first encoder model. Since we know the spans of $e_i$ and $e_j$, it is customary to encapsulate these spans with special tokens~\cite{zhong-chen-2021-frustratingly, ai2023end}. Specifically subject $e_s$ is placed between entity marker tokens \texttt{[SUB:$t_s$]} and \texttt{[/SUB:$t_s$]} to denote the begin and end of the subject entity $e_s$ with entity type $t_s$. Likewise, object $e_o$ is placed between markers \texttt{[OBJ:$t_o$]} and \texttt{[/OBJ:$t_o$]}. The original input along with these demarcated spans is input to the encoder and the output embeddings of the start tokens of $e_s$ and $e_o$ are concatenated to represent the candidate pair. With $\mathcal{E}_{T}$ denoting the input encoder, the associated input representation is
\begin{align}
     \rho_T(e_s, e_o) = \mathbf{W_T}(\mathcal{E}_{T} \texttt{[SUB:$t_s$]} \,  \Vert 
     \, \mathcal{E}_{T} \texttt{[OBJ:$t_o$]}), 
\end{align}
where the concatenated embedding is subjected to a linear transformation $\mathbf{W_T}$. The entity markers are crucial, given that it is important to capture the roles of subjects and objects and their types in determining viable predicates informed by contextual cues. 

\textbf{\textit{Predicate description representations}: }
\label{sec-pred}
We recall that predicates in scientific areas have official descriptions of what they are expected to capture. For example, in BioRED dataset~\cite{luo2022biored}, the \texttt{POSITIVE CORRELATION} relation between \texttt{Chemical} and \texttt{Disease} entities is described as:  ``The drug A may induce the disease B, increase its risk, or the levels may correlate with disease risk.'' For the \texttt{NULL} predicate, we simply describe it as: ``There are no relations between the drug A and disease B.'' When we find that the original definitions are overly simplistic or not sufficiently informative, we make necessary modifications to enhance clarity  by prompting GPT-4. For example, in the SciERC dataset~\cite{luan2018multitask}, the original definition provided for the predicate \texttt{PART-OF} reads:  ``B is a part of A.'' This definition, while broad, lacks sufficient elaboration and specificity needed for  model training. To address this, we revise the definition to better capture the essence of the relationship, making it more informative and directly applicable for our purposes. Our revised definition, crafted to enhance clarity, is: ``B is a component or segment that is integral to the structure or composition of A.''

\begin{figure}[t!]
    \centering
    \includegraphics[scale=0.5]{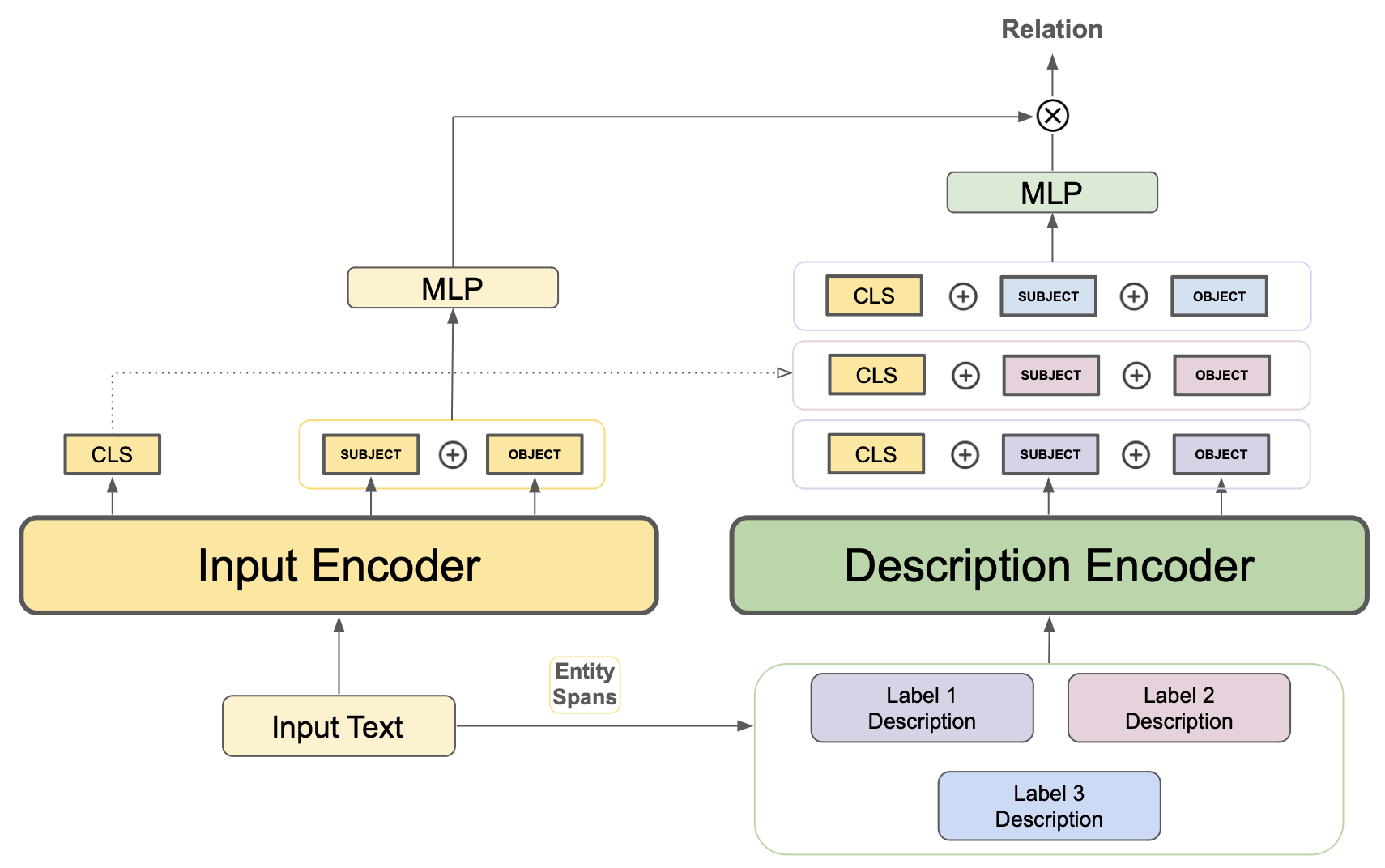}
    \caption{Our model architecture in detail with a 3-class example: we incorporate instance information by inserting entity spans from the input text and concatenate the \texttt{[CLS]} embedding or the input encoder to the description encoder representations. (Note that the linear and softmax layers used during training for equation \eqref{eq-ce} are not shown here.) 
    %\textcolor{blue}{As the description inputs are quite short, the description encoder only requires minimal computational overhead. (95\% FLOPs saved compared to full size inputs.)
    }
    %traditional methods uses a fixed classifier, our description encoder computes instance-specific embeddings for the final relation prediction.
    
    \label{fig:main-architecture}
\end{figure}

Instance adaption is accomplished by inserting the entity spans from the input into natural place holders for each predicate description. The main rationale for adaptation is to encode the entity spans in the context of the language used in the canonical definitions rather than simply using the definition without grounding in specific entities used. Since identifying subject/object placeholder slots in definitions is a one-time task for each predicate, this could be done manually. Entity spans from the input are directly inserted into the chosen placeholder slots to create instance specific predicate descriptions, as shown for the \texttt{TREATS} predicate in the Introduction. 
Though incorporating entity spans in the description texts grounds their representation, it primarily focuses on the ``hard tokens'', which may not capture the full essence of contextual nuances present in the input text. To address this potential limitation, we also incorporate the \texttt{[CLS]} representation from the first encoder \(\mathcal{E}_{T}\) into the
predicate description representation. 
Thus the instance-adapted representation for a specific $r \in R$ is
\begin{align}
    \rho^r_{D} (e_s, e_o) = \mathbf{W_D}(\mathcal{E}_{T}[\texttt{CLS}] \,\Vert\, 
    \mathcal{E}^r_{D} 
    \texttt{[SUB:$t_s$]} \Vert &\mathcal{E}^r_{D} \texttt{[OBJ:$t_o$]}),
\end{align}
where $\mathcal{E}_D$ is description encoder,  $\mathcal{E}^r_D$ is the representation derived for description of predicate $r$ grounded with entity spans from the input, and $\mathbf{W_D}$ is a linear transformation. By matching projection dimensions for  $\mathbf{W_T}$ and $\mathbf{W_D}$, $\rho_T$ and $\rho^r_D$ lend themselves to similarity comparisons. Given the description texts are typically short sentences (usually less than 20 words), the description encoder is substantially cheaper to run compared to the input encoder; the small number of its input tokens leads to only $\approx$ 5\% FLOPs of the computation that a full 512-token model requires. Hence, this does not significantly raise the training or inference cost in the dual-encoder configuration.

\textit{Remark:} As we indicated earlier, biomedicine is a field where all predicates have well understood descriptions often written by experts in the field. While this would be true for any scientific discipline, in the off chance they are not available and there is a need to generate descriptions on the fly along with automatic insertion of placeholders, we came up with a simple prompt to generate them: ``Explain the $\{rel\}$ relationship between a $\{e1_{type}\}$ (subject) and a $\{e2_{type}\}$ (object) by crafting a sentence that includes the placeholders @subject@ and @object@''. When instantiating this prompt template with ``treats'' ($\{rel\}$), ``drug'' ($\{e1_{type}\}$), and ``disease'' ($\{e2_{type}\}$), GPT-4o generated the description: \textit{@subject@ is prescribed for the treatment of @object@, helping to manage symptoms and improve patient outcomes by targeting the disease mechanisms.} A second run generated this slightly different description: \textit{@subject@ treats @object@ by targeting the underlying causes or symptoms of the disease, aiding in recovery or management}. We provide this as part of the code in the \texttt{generate\_descriptions.py} file in the GitHub repo provided in the Introduction. That said, we do not recommend using LLM generated predicate descriptions, especially if the predicate is polysemous or homophonic, in which case the LLM generation may not align with user intention.  

\textbf{\textit{Contrastive and cross entropy objectives}: }
\label{sec-loss}
Although two entities can be linked via multiple predicates in the real world, for a specific input textual instance, it is generally safe to assume only one predicate is at play. Using this multiclass (and not multilabel) assumption, we formulate a contrastive objective to push the representations of $\rho_T$ and $\rho^{r^+}_D$ ($r^+ \in R$, the correct predicate) closer to each other while pushing $\rho^{r^-}_D$ away from $\rho_T$ for all $r^- \in R \setminus \{r^+\}$, the incorrect predicates. We represent this closeness/farness via vector similarity $sim (\rho_T, \rho^r_D) = \cos (\rho_T,\rho^r_D)$. We chose the cosine distance which is naturally in the [0, 1] range as it was better than the dot product, which produced suboptimal performance due to scaling issues in initial experiments. (Note that we still use the normalized dot product formulation for cosine implementation instead of calling Python's math.cos().) With this setup, the contrastive loss function for a given input using the instance adapted predicate descriptions is
\begin{eqnarray}
\label{eq-ct}
 L_{ct}((e_s, e_o), r^+, r^-_1, \ldots, r^-_{|R|-1}) 
= -\log \frac{ e^{\mathrm{sim}(\rho_T, \rho^{r^+}_D)} }{e^{\mathrm{sim}(\rho_T, \rho^{r^+}_D)} + \sum_{j}{e^{\mathrm{sim}(\rho_T, \rho^{r^-_j}_D)}}}
\end{eqnarray}

During our implementation, though contrastive training (from Equation \eqref{eq-ct}) was effective in pushing the input and positive predicate representations closer, we observed it does not learn robust relation representations of the input (the $\rho_T$s). To address this, we propose to  add a new linear layer and simultaneously optimize the cross-entropy  loss 
\begin{eqnarray}
\label{eq-ce}
    L_{ce} (e_s, e_o) = -\sum_{r \in R}  y_{r} \log(p_{r})
\end{eqnarray}
where 
$y_r$ is a binary indicator of whether class label $r$ is the correct classification for the instance and $p_r$ is the predicted probability of the instance belonging to class $r$.
We use the unified loss 
\begin{eqnarray}
\label{eq-final}
    L_u = \alpha L_{ce} + (1 - \alpha) L_{ct}
\end{eqnarray}
during training,
where the $ 0 \leq \alpha \leq 1$  serves as a hyper-parameter that determines the  influence of the contrastive loss component in the overall loss. Although training is done via Equation \eqref{eq-final}, our model exclusively relies on the contrastive scores to make predictions at inference time as 
\[ r_{pred} = \operatorname*{argmax}_{r \in R} sim(\rho_T, \rho^r_D). \]
The full architecture is shown in Figure~\ref{fig:main-architecture} with the two encoders handling the input and instance-specific label descriptions separately.

\textbf{\textit{Datasets}: }
We looked for public RE datasets that encompass a variety of relation types with apt predicate definitions and landed on four: SciERC, Semeval 2010 task-8, ChemProt, and BioRED with statistics as shown in Table~\ref{tab:dataset}.

\begin{table}[!ht]
    \centering
    \caption{Statistics of datasets used (columns 3--5 are numbers of abstracts/sentences, not relations).}
    \small
\renewcommand{\arraystretch}{1.3}
    \begin{tabular}{lcrrr}
      \toprule
      Dataset & \# Predicates & \# Train & \# Dev & \# Test \\
      \midrule
      SciERC   & 7 & 350 & 50 & 100 \\
      Semeval & 9 & 6507 & 1493 & 2707 \\
      ChemProt & 5/13 & 1020 & 612 & 800\\
      BioRED & 8 & 400 & 100 & 100\\
      \bottomrule
    \end{tabular}
    \label{tab:dataset}
\end{table}

\begin{itemize}
    \item \textbf{SciERC} \cite{luan2018multitask}: This dataset is created from AI conference or workshop paper abstracts and includes annotations for both entities and relationships offering predicates common in scientific discourse. 
    \item \textbf{ChemProt} \cite{krallinger2017overview}: This dataset is designed for chemical-protein interaction detection in biomedical literature and was created as part of the BioCreative shared task series. 
    %It involves the identification of relationships between chemical compounds and proteins, which are vital for tasks such as drug discovery and understanding metabolic pathways.
    %The dataset's focus on a narrow area of biomedicine allows us to assess the precision of our model in a highly specialized context. 
    %The ChemProt corpus encompasses a total of 4,966 abstracts, which are distributed into 1,020 documents for training, 612 for development, and 800 for testing. This structured division ensures a balanced approach to training, validating, and evaluating the performance of relation extraction models specifically tailored to biomedical texts.
    %We use a better tokenization methodology proposed by \cite{ai2023endtoend}, which misses only 0.4\% of total entities and 1.37\% of total relations in the training and development datasets. 
    Although entities were annotated with the potential to be connected by one of ten predicates, only five are consistently used for evaluation following the shared task conventions, due to their relative importance and relevance in the context of chemical-protein interactions. These five predicates were further subdivided into a total of 13 fine-grained predicates, which characterize further nuances in interaction types (such as distinguishing between different kinds of upregulators or activators.) We test our methods with both schemes (the five and 13 predicate variations). 
     
    \item \textbf{BioRED} \cite{luo2022biored}: This is a more recent and broad scoped biomedical RE dataset that includes four distinct entity types and eight different predicates. In its original form,   entity normalization  to standardized vocabularies is also expected. We adapt the dataset to fit our needs by treating it as a conventional RE task. In this adaptation, we address the issue of multiple entity mentions associated with a single entity ID by splitting these mentions into separate relations. This modification ensures that each entity mention is treated independently, simplifying the RE process 
     under our current system capabilities. However, it does not reduce   task difficulty because it is evaluated based on obtaining relations between all spans corresponding to both subject and object entities. 

     \item \textbf{SemEval-2010 Task 8} \cite{hendrickx-etal-2010-semeval}: This is a well-known general domain benchmark dataset focused on relation classification. It features nine distinct types of semantic relations such as ``Cause-Effect'' and ``Entity-Origin''. The task was created to evaluate various approaches to the classification of semantic relationships and to offer a standardized testbed for future research efforts.
     
\end{itemize}
SciERC, Semeval and ChemProt deal with sentence-level relations (the participating entities are within the same sentence) but the surrounding context of the full abstract maybe needed to extract the relations. BioRED is a more general document level dataset and includes cross-sentence relations. 

\begin{table*}[!ht]
\centering
\caption{\label{table:SOTA} We compare the micro-$F_1$ score, a common metric for evaluating the accuracy of classification models. Dual-Encoder+Adapt refers to our full model with instance-adaptation. The ChemProt$^{13}$ and ChemProt$^5$ columns refer to the 13-class and 5-class variants of the dataset, respectively.
%Our approach demonstrates an improvement over several state-of-the-art (SOTA) methods, with performance enhancements ranging from 1\% to 2.1\% across four distinct tasks. 
Pretrained biomedical encoders (BioBERT and BiomedBERT) are not used for the general-domain datasets (SciERC and Semeval). ``Multiple'' indicates the use of different encoders for various tasks, as described in the ``base encoder choices'' portion of the Methods section.}
%\small
\renewcommand{\arraystretch}{1.2}
\resizebox{\textwidth}{!}{
\begin{tabular}{lcccccc}
\toprule
\textbf{Methods} & \textbf{Encoder} & SciERC & ChemProt$^{13}$ & ChemProt$^5$ & BioRED & SemEval  \\
\midrule
Devlin et al. \cite{devlin-etal-2019-bert} & BERT\textsc{base} & 65.2 & 68.2 & 73.7 & 34.7 & 89.4 \\
Lee et al. \cite{Lee_2019} & BioBERT & - & 71.8 & 76.5 & 38.7 & - \\
Gu et al. \cite{Gu_2021} & BiomedBERT & - &  72.3 & 77.2  & 48.3 & -\\
Su et al. \cite{su2021improving}  & BiomedBERT & - & - & 78.7 & -  & - \\
GPT-RE \cite{wan2023gptre} & SciBERT  & 69.0 & - & - & -  & 91.9 \\
\midrule
PURE \cite{zhong-chen-2021-frustratingly} & Multiple &  68.5 & 72.5 & 78.7  & 51.4 & 89.9 \\
\midrule
\textbf{Dual-Encoder (\textit{Ours})} & Multiple & 68.6 & 72.6  & 79.5 & 48.9 & 92.0 \\
\textbf{Dual-Encoder+Adapt. (\textit{Ours})} & Multiple & \textbf{71.1} & \textbf{73.5} & \textbf{79.8} & \textbf{52.5} & \textbf{92.7} \\
\bottomrule
\end{tabular}}
\end{table*}

\textbf{\textit{Baseline methods and prior efforts}:  }
We used the basic Google BERT model~\cite{devlin-etal-2019-bert} and its biomedical variants BioBERT~\cite{Lee_2019} and BiomedBERT~\cite{Gu_2021} (also known as PubMedBERT), trained on PubMed corpora, as our baselines. While BioBERT uses the original BERT tokenizer, BiomedBERT's vocabulary is built from scratch and has shown improvements in the past over BERT and BioBERT. Another recent popular method that revived pipelines by using entity role and type specific markers is the Princeton University Relation Extraction (PURE) framework~\cite{zhong-chen-2021-frustratingly}. PURE uses a BERT model as its base and builds on it with special tokens for entity boundaries. 
We also compare with two other prior efforts. The first by \cite{su2021improving} introduced a novel method that enhances RE capabilities with contrastive learning for data augmentation. This approach refines text representations derived from the BERT model specifically for RE tasks. \cite{wan2023gptre} developed GPT-RE, a new RE system that integrates GPT-3, using an LLM as an instance-aware retrieval mechanism to obtain relevant demonstrations from the training dataset. The demonstrations are then used for in-context learning to predict outputs with GPT-3.

\textbf{\textit{Base encoder choices and settings}: } 
\label{sec-bs}
Our method relies on two encoder models working together. 
Following prior studies \cite{zhong-chen-2021-frustratingly, wan2023gptre}, we use SciBERT \cite{beltagy2019scibert} as the encoder for the AI relation extraction dataset SciERC. SciBERT has been pre-trained on a corpus of computer science and biomedical full text articles, which makes SciBERT  well-suited for the SciERC dataset.
We use a general-domain encoder model BERT-base \cite{devlin-etal-2019-bert} for Semeval dataset followed by prior study \cite{wan2023gptre}.
For BioRED and ChemProt, we use BiomedBERT  as the encoder. 
%BiomedBERT is a variant of BERT that has been specifically pre-trained on a large corpus of biomedical literature extracted from PubMed abstracts. This specialization makes BiomedBERT well-suited for tasks involving biomedical text, as it is trained to understand the unique vocabulary and complex structures found in this domain. We use those selections for both our baseline models and our proprietary models to ensure a fair and consistent comparison of performance. 
In all our experiments, the encoders are all the same size as BERT-base, which contains approximately 110 million parameters.

All of our experiments were conducted using a consistent training regimen across different models. We arrived at $\alpha =0.5 $ with experiments on the development datasets for weighting the two losses in Equation~\eqref{eq-final}.
Each model was trained for 10 epochs, a batch size of 4 with a single run. We experimented with learning rates 1e-5 and 2e-5, to optimize performance and adaptability across various tasks and datasets. We used one NVIDIA V100 GPU trained for roughly 10 hours per experiment.

\section*{Results and Discussion}
\vspace{-1mm}
\textbf{\textit{Main results}: }
%We conduct a thorough comparison of our main results against a variety of previous studies \cite{Gu_2021, su2021improving, wan2023gptre} as well as our established baseline model, PURE \cite{zhong-chen-2021-frustratingly}. To provide a comprehensive evaluation, we analyze and report the performance scores across the entire test sets of four distinct tasks. 
The main results in Table \ref{table:SOTA} show our method (last row) leads to performance enhancements ranging from 1\% to 2.1\% in micro F-score over prior methods, observed both in biomedical and general domain settings. Since BioRED was introduced after papers from the first three rows were published, we trained and ran their code on it; this was also done for the 13-class version of ChemProt. Since the paradigm used by \cite{su2021improving} (augmentation) and 
\cite{wan2023gptre} (GPT-3 calls) were quite different from ours, we did not run new experiments with them and simply reported the results from their papers, when available. 
In the penultimate row, we show the scores from a standard dual-encoder model without instance-adaptation. That is, we simply used the static predicate definitions without instantiating them with the entities from the input. The instance specific version shows nontrivial gains of 2.5\% on the SciERC dataset and 3.6\% on the BioRED dataset. This shows that instance-adaptation features are crucial for this method.
Additionally, our experiments with the BioRED dataset were conducted at the mention-level, rather than at the entity identifier level. This evaluation choice was driven by our focus on the granularity of mention-specific data rather than on broader entity identifiers, dealing with which is an orthogonal issue of entity linking. %Consequently, our results from the BioRED dataset are not directly comparable to those methodologies that perform entity normalization. 

\textbf{\textit{Ablation of the \texttt{[CLS]} component}: }
Recall that the first input encoder's \texttt{[CLS]} token output was included as part of the  predicate description representation  to enhance its instance specific aspects, as discussed in the ``Predicate description representations'' portion of the Methods section. In the 2nd row of  Table~\ref{tb:ablation}, we see that removing this component dips the performance by 0.7 points in F-score for the ChemProt dataset indicating a modest influence on eventual performance. 
\begin{table}[h]
\centering
\caption{\texttt{ChemProt (5-class)} ablated $F_1$ scores on the test set.}
%\small
 \renewcommand{\arraystretch}{1.2}
% \resizebox{0.47\textwidth}{!}{
\begin{tabular}{lccc}
\toprule
\textbf{Model} & \texttt{ ChemProt$^5$} $F_1$ \\\midrule
\textbf{Full dual-encoder model}  & \textbf{79.8} \\
 \,\, w/o. \texttt{[CLS]} concatenation   & 79.1 \\
 \,\, w/o. cross-entropy loss & 78.7 \\
 \,\, w/o. dual-encoder & 78.5 \\
\bottomrule
\end{tabular}
\label{tb:ablation}
\end{table}

\textbf{\textit{Ablation of cross-entropy loss}: }
\label{sec-ce}
Integrating cross-entropy loss alongside our contrastive loss proved helpful for enhancing the model's ability to establish more effective relation representations. Traditional approaches in RE have relied on cross-entropy loss due to its effectiveness in clustering input embeddings of the same class closely together,  improving overall model performance. From row 3 of Table~\ref{tb:ablation}, we see that dropping it leads to a 1.1\% dip in $F_1$ score. 

We examined the impact of cross-entropy loss on overall model performance by varying its weight in the combined loss function (Equation~\eqref{eq-final}) to determine and fix its value for final training as part of hyperparameter optimization. The results are presented in Table \ref{tab:celoss} for the ChemProt development dataset. Assigning a small weight to this loss ($\alpha=0.1$) and hence using more of the contrastive component leads to the highest recall; but the highest F-score is reached when $\alpha=0.5$ (equal weighting), with some compromise in recall but with a better jump in precision sufficient enough to lead to an overall $F_1$ gain of 1\%.

\begin{table}[ht!]
    \centering
    \caption{Different proportions of cross-entropy loss on ChemProt (5-class) development set.}
 %   \small
    \renewcommand{\arraystretch}{1.2}
    \begin{tabular}{cccc}
    \toprule
      $L_{ce}$ weight  & $P$ & $R$ & $F_1$ \\
      \midrule
        $\alpha=0.1$ & 76.1 & 87.1 & 81.3 \\
        $\alpha=0.3$ & 81.2 & 82.8 & 81.9 \\
        $\alpha=0.5$ & 82.9 & 81.8 & 82.3 \\
        $\alpha=0.7$ & 80.9 & 83.2 & 82.1 \\
    \bottomrule
    \end{tabular}
    \label{tab:celoss}
\end{table}

\textbf{\textit{Ablation of the dual encoder}: }
Instead of the two separate encoders for input text and predicate description representation, we used the same encoder (hence shared parameters). 
Row 4 of Table~\ref{tb:ablation} shows that this results in a 1.3\% dip in performance.
This maybe due to potential trade-off between system efficiency and performance. The dual-encoder configuration seems to provide superior performance by leveraging specialized processing streams for input and description texts.

\textbf{\textit{Assessing different descriptions}: }
Although we used predicate definitions created by RE dataset creators, they are nevertheless a single symbolic and discrete form. 
We wondered about how the scores would change if we rephrased them by prompting GPT-4 to ``rewrite'' them without any special instructions. 
We also tested a simple baseline that just has the predicate name and entity spans fill the slots in this template: \texttt{@predicate@: @subject@, @object@}.

\begin{table}[!h]
\centering
\caption{\texttt{ChemProt (13-class)} scores with different description texts. }
 \renewcommand{\arraystretch}{1.2}
% \resizebox{0.47\textwidth}{!}{
\begin{tabular}{lccc}
\toprule
\textbf{Versions} & \texttt{ChemProt$^{13}$} $F_1$ \\\midrule

\textsc{Original}   & 73.5 \\
\textsc{Rewrite}  & 73.4 \\
\textsc{Simple} & 72.9 \\
\bottomrule
\end{tabular}
\label{tb:description}
\end{table}

As the  associated scores in Table~\ref{tb:description} show, there is not much difference in performance with rewrites. However, surprisingly, the simple baseline is worse than the original definition by only $0.6\%$. This small dip highlights the effectiveness of even using templated forms with the tokens indicating the predicate name. But it is important to 
note that predicate names in ChemProt dataset are highly specific with unambiguous meanings such as \texttt{UPREGULATOR} and \texttt{ANTAGONIST}. Pretrained encoders such as BiomedBERT might already have decent representations for them that carry substantial semantic signal. 
However, it is not clear if the simple baseline holds as well with predicate names that have a broader meaning, where explicit detailed definitions might be needed for more gains. 

{\textbf{\textit{Pipeline setup with an NER tool for end-to-end evaluation}: }
Without assuming the availability of gold entity spans, we evaluate our model in an end-to-end manner using a pipeline that connects an NER tool's output to our relation classifier. Specifically, we first apply the span-based PURE model~\cite{ai2023end} to identify entities, and then use these predictions as inputs to our relation extraction model (Figure~\ref{fig:main-architecture}). This reflects a realistic  scenario enabling us to assess the model’s robustness under practical conditions. Since relation extraction depends on accurate entity boundaries and types, errors from the NER stage may propagate and affect overall performance. By adopting the PURE NER component, a strong baseline aligned with prior work, we ensure a fair evaluation. The comparison results in this pipeline setting are presented in Table~\ref{tab:pipeline}. Our model, when combined with the PURE NER component, achieves state-of-the-art performance, surpassing previous approaches in the end-to-end setting. This demonstrates that the pipeline setup, despite relying on predicted entities, remains highly effective and outperforms existing joint models~\cite{ zuo2022span,sun2024joint}.

\begin{table}[]
    \centering
    \caption{End-to-end evaluation of our model in a pipeline setup on \texttt{ChemProt$^{5}$}, the 5-class variant of the dataset}
    \renewcommand{\arraystretch}{1.2}
    \begin{tabular}{lccc}
    \toprule
    \multirow{2}{*}{\textbf{Model}} & \multicolumn{3}{c}{End-to-end RE} \\
    \cmidrule(lr){2-4}
     & $P$ & $R$ & $F_1$ \\
    \midrule
    SpanMB\_BERT \cite{zuo2022span} & 68.0 & 61.5 & 64.6 \\
    MCTPL~\cite{sun2024joint}  & 64.7 & 66.7 & 65.7 \\
    PURE~\cite{ai2023end}  & 70.8 & 67.2 & 69.0 \\
    \textbf{Ours} & 72.5 & 68.5 & \textbf{70.4} \\
    \bottomrule
    \end{tabular}
    \label{tab:pipeline}
\end{table}

\section*{Error Analysis}
\vspace{-1mm}
Although our approach does not incorporate the NER step, there are some notable errors purely for the RE component. One issue is the model's failure to accurately infer the meaning from complex or ambiguous contexts. 

Consider this example from the SciERC dataset: 
\textit{``The proposed detectors are able to capture large-scale structures and distinctive textured patterns, and exhibit strong invariance to rotation, illumination variation, and blur.''} 
While our model successfully identified the \texttt{USED-FOR} relation between  \textit{detectors} (subject) and entities \textit{large-scale structures} and \textit{distinctive textured patterns} (object), it was not able to predict the same gold \texttt{USED-FOR}  relation with object entities \textit{rotation}, \textit{illumination variation} and \textit{blur}. Since the sentence does not explicitly state that the \textit{detectors} are used for 
\textit{rotation}, \textit{illumination variation} and \textit{blur}, it might have missed potential implied links.  On the other hand, one could argue from the original input that maybe the ground truth \textit{detectors} \texttt{USED-FOR} relations with these entities may not be entirely accurate --- \textit{detectors} are overcoming the barriers of \textit{blur} and \textit{rotation} to excel at capturing \textit{large-scale structures} and \textit{distinctive textured patterns} and not necessarily being used to detect or capture blur/rotation.  

Next, consider this input from the ChemProt dataset: \textit{``Down-regulation of prostate-specific antigen (PSA) expression, an AR-target gene, by estramustine and bicalutamide was accompanied by the blockade of the mutated androgen receptor.''}
The model was able to identify the \texttt{INDIRECT-DOWNREGULATOR} relation between drugs \textit{estramustine} and \textit{bicalutamide} (subjects) and the protein \textit{PSA} (object). But it failed to spot the same relation of those drugs with the object protein \textit{AR}, which appears to be implicitly stated. Considering ChemProt extraction sometimes involves the full abstract, other sentences surrounding this may offer indirect clues about the relation. However, upon examining the full abstract, we are not able to see stronger evidence than what is already present in the sentence shown here.

Another common error pattern is due to the model's lack of grasp of deep domain knowledge, particularly in biomedical datasets when information is densely packed. For example, consider this pithy ChemProt input:
\textit{``In vivo, agonist actions of yohimbine at 5-HT(1A) sites are revealed by WAY 100,635-reversible induction of hypothermia in the rat.''} Here the gold relation is the \texttt{ANTAGONIST} link between subject \textit{WAY 100,635} and object \textit{5-HT(1A)}. The agonist actions of the chemical yohimbine on 5-HT(1A) result in hypothermia and the fact that this hypothermia can be reversed by \textit{WAY 100,635} indicates that it is playing an antagonist role for \textit{5-HT(1A)}. This is a complex expression involving the intermediate entity yohimbine and an unusual looking phrase \textit{WAY 100,635-reversible} that densely packs meaning that typically needs a new sentence to convey explicitly. This may simply be a case of a highly complex example needing multi-hop reasoning that needed to be carried out with a compact input text. 

A significant challenge in the BioRED dataset arises from the hierarchical nature of predicates where a more general \texttt{ASSOCIATION} relation is confused with specific \texttt{POSITIVE/NEGATIVE CORRELATION} relations.  If the model fails to latch on to specific relations, it may default to the general predicate. Consider the input:
\textit{``The 2:1 atrioventricular block improved to 1:1 conduction only after intravenous lidocaine infusion or a high dose of mexiletine, which also controlled the ventricular tachycardia. A novel, spontaneous LQTS-3 mutation was identified in the transmembrane segment 6 of domain IV of the Na(v)1.5 cardiac sodium channel, with a G$\rightarrow$A substitution at codon 1763, which changed a valine (GTG) to a methionine (ATG). \ldots.''}
Here the text mentions that the administration of the drug \textit{lidocaine} improved the patient's condition, and separately, a novel mutation (\textit{LQTS-3}) was identified in patients suffering from arrhythmias. The gold relation is \texttt{NEGATIVE CORRELATION} of \textit{lidocaine} with \textit{LQTS-3 mutation} but the model predicted \texttt{ASSOCIATION}, the generic predicate, which is only supposed to be used when the correlation type cannot be discerned from the input. However, although \textit{lidocaine} and \textit{LQTS-3} are never mentioned in the same sentence, through the full abstract that has over 250 words, one can see the \textit{lidocaine} is blocking the function of this mutation in causing arrhythmias and as such has a negative correlation with it.

\section*{Conclusion}
\vspace{-2mm}
In this paper, we introduce a new approach to RE that uses a dual-encoder architecture leveraging predicate descriptions that compares input text to instance-adapted canonical descriptions of the predicates. Experiments with an equally weighted joint contrastive and cross-entropy loss show that this approach improves over prior methods for four RE datasets involving biomedical and general domain tasks. Ablation experiments also reveal that each component of the model plays a nontrivial role in the overall performance. We conclude with two future research directions.
\begin{itemize}
    \item As discussed in the Methods section, we only use the cosine scores of the two encoder representations to pick the right predicate at test time. Since  cross entropy loss helped during training, a better way to integrate output softmax probability estimates with normalized cosine scores could lead to more performance improvements.
    
    \item Many RE use-cases have hierarchical predicate structures which were also observed in ChemProt and BioRED datasets in our paper. More involved learning strategies that leverage label hierarchies, potentially with graph convolutional nets, may be needed. Another training loss term that imposes penalties for violating hierarchical constraints could lead to better regularization and fewer errors arising from distant predicates from the hierarchy.
\end{itemize}

% Despite the overall positive results, our model has some limitations. 
% % Although the description texts are derived from the annotation guidelines, they require an additional step to make them compatible with our model --- insertion of placeholders for the subject and object entities. However, this is a one-time pre-procesing step. Also, 
% the instance adaptation provided by our current approach is limited to canonical definitions. But, this issue could be mitigated by using LLMs to produce more refined descriptions.
% We conclude with a few future research directions.
% \begin{itemize}
%     \item As discussed in Section~\ref{sec-loss}, we only use the cosine scores of the two encoder representations to pick the right predicate at test time. Since cross entropy loss helped during training, a better way to integrate output softmax probability estimates with normalized cosine scores could lead to more performance improvements.
    
%     \item Many RE use-cases have hierarchical predicate structures which were also observed in ChemProt and BioRED datasets in our paper. More involved learning strategies that leverage label hierarchies, potentially with graph convolutional nets, may be needed. Another training loss term that imposes penalties for violating hierarchical constraints could lead to better regularization and fewer errors arising from distant predicates from the hierarchy.
    
% \end{itemize}

\section*{Acknowledgments} 
\vspace{-2mm}
This project has been funded through National Library of Medicine via grant R01LM013240.
The content is solely the responsibility of the authors and does not necessarily represent the official
views of the NIH.

% References as numbers
\makeatletter
\renewcommand{\@biblabel}[1]{\hfill #1.}
\renewcommand{\bibsection}{%
  \section*{\centering \refname}%
}
\makeatother

% unstr is used to keep citation order
\bibliographystyle{vancouver}
\bibliography{amia}  

\begin{thebibliography}{10}

\bibitem{riloff1993automatically}
Riloff E, et~al.
\newblock Automatically constructing a dictionary for information extraction tasks.
\newblock In: AAAI. vol.~1. Citeseer; 1993. p. 2-1.

\bibitem{zelenko-etal-2002-kernel}
Zelenko D, Aone C, Richardella A.
\newblock Kernel Methods for Relation Extraction.
\newblock In: Proceedings of the 2002 Conference on Empirical Methods in Natural Language Processing ({EMNLP} 2002); 2002. p. 71-8.

\bibitem{bunescu-mooney-2005-shortest}
Bunescu R, Mooney R.
\newblock A Shortest Path Dependency Kernel for Relation Extraction.
\newblock In: Proceedings of Human Language Technology Conference and Conference on Empirical Methods in Natural Language Processing; 2005. p. 724-31.

\bibitem{kambhatla2004combining}
Kambhatla N.
\newblock Combining lexical, syntactic, and semantic features with maximum entropy models for information extraction.
\newblock In: Proceedings of the ACL interactive poster and demonstration sessions; 2004. p. 178-81.

\bibitem{nguyen2015relation}
Nguyen TH, Grishman R.
\newblock Relation extraction: Perspective from convolutional neural networks.
\newblock In: Proceedings of the 1st workshop on vector space modeling for natural language processing; 2015. p. 39-48.

\bibitem{miwa2016end}
Miwa M, Bansal M.
\newblock End-to-End Relation Extraction using LSTMs on Sequences and Tree Structures.
\newblock In: Proceedings of the 54th Annual Meeting of the Association for Computational Linguistics (Volume 1: Long Papers); 2016. p. 1105-16.

\bibitem{vu2016combining}
Vu NT, Adel H, Gupta P, Sch{\"u}tze H.
\newblock Combining Recurrent and Convolutional Neural Networks for Relation Classification.
\newblock In: Proceedings of the 2016 Conference of the NAACL: HLT; 2016. p. 534-9.

\bibitem{guo2019attention}
Guo Z, Zhang Y, Lu W.
\newblock Attention Guided Graph Convolutional Networks for Relation Extraction.
\newblock In: Proceedings of the 57th Annual Meeting of the ACL; 2019. p. 241-51.

\bibitem{lin2019bert}
Lin C, Miller T, Dligach D, Bethard S, Savova G.
\newblock A BERT-based universal model for both within-and cross-sentence clinical temporal relation extraction.
\newblock In: Proceedings of the 2nd Clinical Natural Language Processing Workshop; 2019. p. 65-71.

\bibitem{joshi2020spanbert}
Joshi M, Chen D, Liu Y, Weld DS, Zettlemoyer L, Levy O.
\newblock Spanbert: Improving pre-training by representing and predicting spans.
\newblock Transactions of the association for computational linguistics. 2020;8:64-77.

\bibitem{zhong-chen-2021-frustratingly}
Zhong Z, Chen D.
\newblock A Frustratingly Easy Approach for Entity and Relation Extraction.
\newblock In: Proceedings of the 2021 Conference of the NAACL: HLT; 2021. p. 50-61.

\bibitem{li2023revisiting}
Li G, Wang P, Ke W.
\newblock Revisiting Large Language Models as Zero-shot Relation Extractors.
\newblock In: Findings of the Association for Computational Linguistics: EMNLP 2023; 2023. p. 6877-92.

\bibitem{wadhwa2023revisiting}
Wadhwa S, Amir S, Wallace BC.
\newblock Revisiting Relation Extraction in the era of Large Language Models.
\newblock In: Proceedings of the 61st Annual Meeting of the ACL; 2023. p. 15566-89.

\bibitem{ai2023end}
Ai X, Kavuluru R.
\newblock End-to-End Models for Chemical--Protein Interaction Extraction: Better Tokenization and Span-Based Pipeline Strategies.
\newblock In: 2023 IEEE 11th International Conference on Healthcare Informatics (ICHI). IEEE; 2023. p. 610-8.

\bibitem{luo2022biored}
Luo L, Lai PT, Wei CH, Arighi CN, Lu Z.
\newblock BioRED: a rich biomedical relation extraction dataset.
\newblock Briefings in Bioinformatics. 2022;23(5):bbac282.

\bibitem{luan2018multitask}
Luan Y, He L, Ostendorf M, Hajishirzi H.
\newblock Multi-Task Identification of Entities, Relations, and Coreferencefor Scientific Knowledge Graph Construction.
\newblock In: Proceedings of the EMNLP; 2018. p. 3219–-3232.

\bibitem{krallinger2017overview}
Krallinger M, Rabal O, Akhondi SA, P{\'e}rez MP, Santamar{\'\i}a J, Rodr{\'\i}guez GP, et~al.
\newblock Overview of the BioCreative VI chemical-protein interaction Track.
\newblock In: Proceedings of the sixth BioCreative challenge evaluation workshop. vol.~1; 2017. p. 141-6.

\bibitem{hendrickx-etal-2010-semeval}
Hendrickx I, Kim SN, Kozareva Z, Nakov P, {\'O}~S{\'e}aghdha D, Pad{\'o} S, et~al.
\newblock {S}em{E}val-2010 Task 8: Multi-Way Classification of Semantic Relations between Pairs of Nominals.
\newblock In: Proceedings of the 5th International Workshop on Semantic Evaluation; 2010. p. 33-8.

\bibitem{devlin-etal-2019-bert}
Devlin J, Chang MW, Lee K, Toutanova K.
\newblock {BERT}: Pre-training of Deep Bidirectional Transformers for Language Understanding.
\newblock In: Burstein J, Doran C, Solorio T, editors. Proceedings of the 2019 Conference of the NAACL: HLT; 2019. p. 4171-86.

\bibitem{Lee_2019}
Lee J, Yoon W, Kim S, Kim D, Kim S, So CH, et~al.
\newblock BioBERT: a pre-trained biomedical language representation model for biomedical text mining.
\newblock Bioinformatics. 2019 Sep;36(4):1234–1240.

\bibitem{Gu_2021}
Gu Y, Tinn R, Cheng H, Lucas M, Usuyama N, Liu X, et~al.
\newblock Domain-Specific Language Model Pretraining for Biomedical Natural Language Processing.
\newblock ACM Transactions on Computing for Healthcare. 2021 Oct;3(1):1–23.

\bibitem{su2021improving}
Su P, Peng Y, Vijay-Shanker K.
\newblock Improving BERT Model Using Contrastive Learning for Biomedical Relation Extraction.
\newblock In: Proceedings of the 20th Workshop on Biomedical Language Processing; 2021. p. 1-10.

\bibitem{wan2023gptre}
Wan Z, Cheng F, Mao Z, Liu Q, Song H, Li J, et~al.
\newblock GPT-RE: In-context Learning for Relation Extraction using Large Language Models.
\newblock In: Proceedings of the EMNLP; 2023. p. 3534-47.

\bibitem{beltagy2019scibert}
Beltagy I, Lo K, Cohan A.
\newblock SciBERT: A Pretrained Language Model for Scientific Text.
\newblock In: Proceedings of the 2019 Conference on EMNLP and the 9th IJCNLP; 2019. p. 3615-20.

\bibitem{zuo2022span}
Zuo M, Zhang Y.
\newblock A span-based joint model for extracting entities and relations of bacteria biotopes.
\newblock Bioinformatics. 2022;38(1):220-7.

\bibitem{sun2024joint}
Sun Z, Xing L, Zhang L, Cai H, Guo M.
\newblock Joint Biomedical Entity and Relation Extraction Based on Multi-Granularity Convolutional Tokens Pairs of Labeling.
\newblock Computers, Materials \& Continua. 2024;80(3).

\end{thebibliography}

\end{document}